# Text to Insight: Accelerating Organic Materials Knowledge Extraction via Deep Learning

**Xintong Zhao[1], Steven Lopez[2], Semion Saikin[3], Xiaohua Hu[1], Jane Greenberg[1]**
Metadata Research Center, College of Computing and Informatics, Drexel University, USA[1]
Department of Chemistry and Chemical Biology, College of Science, Northeastern University, USA[2]
Kebotix, Inc., USA[3]
{xz485, xh29, jg3243}@drexel.edu[1], {s.lopez}@northeastern.edu[2], {semion}@kebotix.com[3]

**ABSTRACT**

Scientific literature is one of the most significant resources for sharing knowledge. Researchers turn to scientific literature as a first step in designing an experiment. Given the extensive and growing volume of literature, the common approach of reading and manually extracting knowledge is too time consuming, creating a bottleneck in the research cycle. This challenge spans nearly every scientific domain. For the materials science, experimental data distributed across millions of publications are extremely helpful for predicting materials properties and the design of novel materials. However, only recently researchers have explored computational approaches for knowledge extraction primarily for inorganic materials. This study aims to explore knowledge extraction for organic materials. We built a research dataset composed of 855 annotated and 708,376 unannotated sentences drawn from 92,667 abstracts. We used named-entity-recognition (NER) with BiLSTM-CNN-CRF deep learning model to automatically extract key knowledge from literature. Early-phase results show a high potential for automated knowledge extraction. The paper presents our findings and a framework for supervised knowledge extraction that can be adapted to other scientific domains.

**INTRODUCTION**

Scientific output plays a critical role in sharing and advancing knowledge. Materials science research literature presents a key example, as scientific publications record *material structures*, *properties* and *processing methods* - essentially the core data required for materials design and predictive analysis. The extensive volume of literature, while exciting, also present researchers with a significant challenge. As of April 13th 2021, a count of 9,861,616 materials scientific outputs have been indexed by the Scopus database. Facing millions of documents, manually reading related literature to extract knowledge is simply not a feasible approach for any researcher. Research using computational approaches to extract knowledge from scientific literature has been pursued for inorganic materials although the body of work is fairly limited (e.g., Kim et al., 2017; Weston et al., 2019). Moreover, there is little evidence of computational knowledge extraction focusing on organic materials. This is surprising, given the urgent needs to automate literature analysis across various research domains including chemistry and biomedicine (e.g., Erekhinskaya et al., 2016; Nuzzo et al., 2010; Zhai et al., 2019).

The research presented in this paper takes initial steps to address this gap by pursuing automatic knowledge extraction for organic materials science, drawing from scientific literature. The research reports on the identification of a set of data-like entities and the framework and approach for building a gold standard, annotation activity, and a baseline analysis. The major novelties and contributions include: 1) a knowledge extraction framework that can be applied to different scientific domains; 2) development of NER method with BiLSTM-CNN-CRF model to automate the knowledge extraction for organic materials literature; 3) publication of the source code. Open access to research corpora and datasets is still limited for materials science, particularly compared to other disciplines (Sang & De Meulder, 2003; Kim et al., 2003; Li et al., 2016; Registry of Research Data Repositories). We collected and have built an original data set from the ground-up (e.g., from scratch) and made source code publicly available at https://github.com/noellzhao/organic_mat_knowledge_extraction.

**RELATED WORK**

Motivated by the attention to materials and advances in computational research, researchers have started to look at NLP techniques as a potential solution to the knowledge extraction challenge for materials study. For instance, Tshitoyan et al. (2019) discuss the feasibility of using word embedding, which is an unsupervised approach to convert natural language to vectors of real numbers, to discover latent materials knowledge embedded in massive publications. Their result shows that word embedding could have high potential in inorganic materials discovery. Nevertheless, the work conducted by Shetty and Ramprasad (2021) confirms that word embedding can be helpful for polymer predictions too. Rule-based NLP techniques, such as pattern and string matching have also been discussed. Huang and Cole (2020) apply pattern matching techniques to extract properties specially for battery materials. Kim et al. (2017) use a combination of machine learning and pattern matching methods to predict synthesis parameters.

Materials researchers have studied the needs and compared different methods for materials knowledge extraction. Kim et al. (2019) express the urgent needs to facilitate text-mining accessibility, clarity and reproducibility for accurately extracting knowledge from materials literature. By comparing feature-based and neural network-based methods for extracting synthesis procedures, Mysore et al. (2017) suggest that neural network-based models generally have higher performance than models using hand-engineered features.

A review of research literature shows that studies related to materials science and knowledge extraction are a more recent development, and that the research tends to focus on inorganic materials. Additionally, the published results report using unsupervised/rule-based approaches. Of the research reported, there is very little discussion on automated knowledge extraction for organic materials. This study aims to explore the use of supervised, deep learning methods to extract knowledge for organic materials, which can be also adapted to other scientific domains.

## METHOD

### Research Framework

Figure 1 demonstrates the research framework used to guide this study. Overall, our proposed framework consists of five main components: 1) corpus collection, 2) data annotation, 3) text processing, 4) model training and 5) applications. The first four components are described in the rest of this section, whereas the actual applications part is discussed in the discussion section.

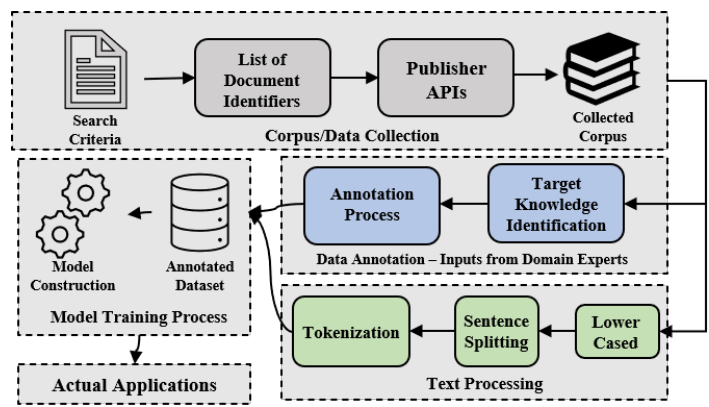

**Figure 1. Research Framework for Automatic Knowledge Extraction using Deep Learning**

### Corpus Collection and Processing

To build our corpus related to organic materials research, we retrieved 92,667 English language scientific literature published between 2000 to 2021 via the Scopus index system. The corpus was collected with keyword search "organic" and "materials science". Retrieved articles are from publishers such as the *Royal Society of Chemistry*, *Springer-Nature* and *American Chemical Society*. We used Scopus associated API to download their abstracts, which overall contain 708,376 sentences (roughly 18 million tokens). All records are lower-cased, then tokenized by the Python library *ChemDataExtractor* (Swain & Cole, 2016) and *NLTK* (Loper & Bird, 2002).

### Data Annotation

A first step for data annotation was to determine target information (entity types) that needed to be extracted. Two scientists with expertise in organic chemistry and organic materials identified the key entity types by first being introduced to the entity used in inorganic materials research (Weston et al., 2019) and then examining the corpus of literature in organic materials to specify relevant entity types. Table 1 presents an initial set of entity types, where definition and examples are also included. Up to now, 855 sentences from the corpus have been manually annotated and stored under IOB (Inside-Outside-Beginning) tagging scheme. Domain expert annotations were reviewed by a third evaluator (an NLP researcher) to address any discrepancies and confirm annotation consistency.

| Entity Type | Definition | Example |
|---|---|---|
| Molecule (MOL) | A fundamental unit of a chemical compound | H2SO4; hydrogen peroxide |
| Polymer (POLY) | A molecule that is majorly composed of multiple similar units | polypyrrole; polyamide |
| Property (PRO) | Fundamental physical or chemical characteristics of a particular compound | optical absorption; binding affinity |
| Characterization Method (CMT) | The method to measure physical or chemical properties | X-ray crystallography; NMR spectroscopy |

**Table 1. Description of Target Entity Types**

**Word Representations**

In this work, both character-level and word-level embeddings are used. We utilized the pre-trained word-level embeddings built by Zhai et al. (2019), which is trained using Word2Vec skip-gram model (Mikolov et al., 2013) with 84,000 unannotated patent documents related to polymer and biomedical science. The character-level embedding is computed using a convolutional (CNN) layer during the training process. At the same time, we are training our word-level embedding based on the collected corpus and FastText (Joulin et al., 2016) model. Since we are still expanding the corpus size, our own word embedding is currently being updated.

**Model Description**

The architecture of Bi-directional LSTM-CNN-CRF model we built for NER is illustrated in Figure 2. While predicting if any tokens belong to the defined knowledge types, the model takes corresponding word-level embedding and char-level embedding computed by CNN layer as input. For each token, its word- and char-embedding are concatenated then fed to the Bi-LSTM (Bi-directional Long Short Term Memory) layer and Conditional Random Field (CRF) output layer accordingly. The model is built using Python library *Keras* (Chollet, 2015), *Tensorflow* (Abadi et al., 2016) and *Scikit-Learn* (Pedregosa et al., 2011).

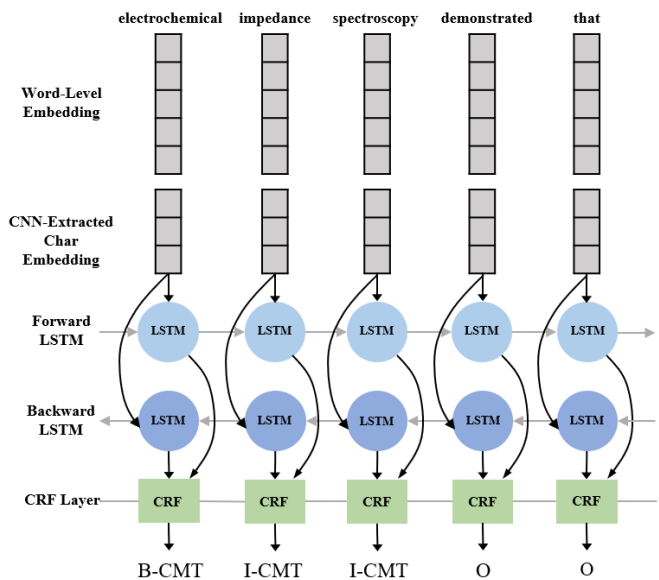

**Figure 2. Illustration of the BiLSTM-CNN-CRF model architecture**

## EXPERIMENT RESULTS

While the collection process is still continuing, we used f-score to evaluate the early-phase performance based on the data we have so far. Table 2 reports the early-phase test result trained on the 684 annotated sentences. We compared the model performance with different embeddings. As shown below (Table 2), the model with our word-level embedding (denoted as "FastText") demonstrates better performance than the embedding trained on patent documents (denoted as "Word2Vec"). The two results using our word embedding seem to have close performance. However, between the two results using embedding trained on patent documents, we can observe that character-level embedding significantly improved the prediction accuracy by 5%-8% for each class.

| Model + Embedding | Molecule (MOL) | Polymer (POLY) | Property (PRO) | Characterization Method (CMT) |
|---|---|---|---|---|
| BiLSTM-CRF+FastText+No Char | **0.67** | 0.67 | **0.56** | 0.74 |
| BiLSTM-CRF+Word2Vec (Zhai et al., 2019)+No Char | 0.58 | 0.56 | 0.49 | 0.49 |
| BiLSTM-CRF+Word2Vec (Zhai et al., 2019)+CNN Char | 0.63 | 0.62 | 0.54 | 0.57 |
| BiLSTM-CRF+FastText+CNN Char | 0.65 | **0.68** | 0.53 | **0.75** |

**Table 2. Test Results from Early-Phase Training Process**

The above test results indicate our approach can have a strong impact on the performance. First, the word embedding trained on closely related corpus can improve the accuracy. The word2vec embedding released by Zhai et al. (2019) are trained on biomedical and chemical corpora, the word2vec embedding encodes the semantic knowledge from patent documents, which could have a different writing style and language than scholarly journal/conference articles. The results confirm that the embedding trained directly on scholarly publications tends to perform better. Second, the training dataset matters. While we gathered more annotated data to the training set, we observed a steady increase in model accuracy. Third, domain expertise is the foundation of scientific knowledge extraction studies and domain experts play a key role in determining the target knowledge and annotating domain-specific corpus. Fourth, char-level embedding can further enhance the performance of the model.

**DISCUSSION AND CONCLUSION**

Recently, the discussion of automated knowledge extraction has expanded across multiple domains. Our research demonstrates the use of supervised learning for knowledge extraction as an approach to address the challenge of manual methods. Additionally, the study helps us to understand what factors can improve the performance of the knowledge extraction model: 1) corpus content; 2) data size; 3) domain expertise and 4) use of embeddings.

Knowledge extraction is gaining attention given the potential in many applications that seek to gain data from unstructured text, such as document retrieval systems. The knowledge extraction process converts unstructured textual data into structured data, and this feature can allow researchers to extract knowledge from large corpuses and store the results as a database. We foresee application as part of our future work.

Finally, we note the current size of our annotated corpus is still comparatively small, although we anticipate higher accuracy as we pursue work to increase the size of our dataset. A final future direction is to expand more entity types: our domain experts analyzed the early-stage training results and suggested that detailed information such as numerical values of processing condition, type of reactions can help researchers to extract more details. Hence, we anticipate to have more entity types added to the model.

**ACKNOWLEDGMENTS**

The research reported on in this paper is supported, in part, by the U.S. National Science Foundation, Office of Advanced Cyberinfrastructure (OAC): Grant: 1940239 and 1940307. We also acknowledge the support of Cyra Gallano and Evan Dubrunfaut from Drexel University, Fatemah Mukadum, Haley Dang and Jordan Cox from Northeastern University, for their role to assist data annotation process.

**REFERENCES**

Abadi, M., Barham, P., Chen, J., Chen, Z., Davis, A., Dean, J., ... & Zheng, X. (2016). Tensorflow: A system for large-scale machine learning. In *12th USENIX Symposium on Operating Systems Design and Implementation (OSDI 16)* (pp. 265-283).

Chollet, F. (2015). Keras documentation. *keras. io*, *33*.

Erekhinskaya, T., Balakrishna, M., Tatu, M., Werner, S., & Moldovan, D. (2016, June). Knowledge extraction for literature review. In *2016 IEEE/ACM Joint Conference on Digital Libraries (JCDL)* (pp. 221-222). IEEE.

Huang, S., & Cole, J. M. (2020). A database of battery materials auto-generated using ChemDataExtractor. *Scientific Data*, *7*(1), 1-13.

Joulin, A., Grave, E., Bojanowski, P., & Mikolov, T. (2016). Bag of tricks for efficient text classification. *arXiv preprint arXiv:1607.01759*.

Kim, E., Huang, K., Kononova, O., Ceder, G., & Olivetti, E. (2019). Distilling a materials synthesis ontology. *Matter*, *1*(1), 8-12.

Kim, E., Huang, K., Saunders, A., McCallum, A., Ceder, G., & Olivetti, E. (2017). Materials synthesis insights from scientific literature via text extraction and machine learning. *Chemistry of Materials*, *29*(21), 9436-9444.

Kim, J. D., Ohta, T., Tateisi, Y., & Tsujii, J. I. (2003). GENIA corpus—a semantically annotated corpus for bio-text mining. *Bioinformatics*, *19*(suppl_1), i180-i182.

Li, J., Sun, Y., Johnson, R. J., Sciaky, D., Wei, C. H., Leaman, R., ... & Lu, Z. (2016). BioCreative V CDR task corpus: a resource for chemical disease relation extraction. *Database*, *2016*.

Loper, E., & Bird, S. (2002). Nltk: The natural language toolkit. *arXiv preprint cs/0205028*.

Mikolov, T., Chen, K., Corrado, G., & Dean, J. (2013). Efficient estimation of word representations in vector space. *arXiv preprint arXiv:1301.3781*.

Mysore, S., Kim, E., Strubell, E., Liu, A., Chang, H. S., Kompella, S., ... & Olivetti, E. (2017). Automatically extracting action graphs from materials science synthesis procedures. *arXiv preprint arXiv:1711.06872*.

Nuzzo, A., Mulas, F., Gabetta, M., Arbustini, E., Zupan, B., Larizza, C., & Bellazzi, R. (2010). Text Mining approaches for automated literature knowledge extraction and representation. *Studies in health technology and informatics*, *160*(Pt 2), 954–958.

Pedregosa, F., Varoquaux, G., Gramfort, A., Michel, V., Thirion, B., Grisel, O., ... & Duchesnay, E. (2011). Scikit-learn: Machine learning in Python. *The Journal of machine Learning research*, *12*, 2825-2830.

Re3data.org - Registry of Research Data Repositories. (n.d.). Retrieved April 19, 2021, from https://www.re3data.org/

Sang, E. F., & De Meulder, F. (2003). Introduction to the CoNLL-2003 shared task: Language-independent named entity recognition. *arXiv preprint cs/0306050*.

Shetty, P., & Ramprasad, R. (2021). Automated knowledge extraction from polymer literature using natural language processing. *Iscience*, *24*(1), 101922.

Swain, M. C., & Cole, J. M. (2016). ChemDataExtractor: a toolkit for automated extraction of chemical information from the scientific literature. *Journal of chemical information and modeling*, *56*(10), 1894-1904

Tshitoyan, V., Dagdelen, J., Weston, L., Dunn, A., Rong, Z., Kononova, O., ... & Jain, A. (2019). Unsupervised word embeddings capture latent knowledge from materials science literature. *Nature*, *571*(7763), 95-98.

Weston, L., Tshitoyan, V., Dagdelen, J., Kononova, O., Trewartha, A., Persson, K. A., ... & Jain, A. (2019). Named entity recognition and normalization applied to large-scale information extraction from the materials science literature. *Journal of chemical information and modeling*, *59*(9), 3692-3702.

Zhai, Z., Nguyen, D. Q., Akhondi, S. A., Thorne, C., Druckenbrodt, C., Cohn, T., ... & Verspoor, K. (2019). Improving chemical named entity recognition in patents with contextualized word embeddings. *arXiv preprint arXiv:1907.02679*